 \documentclass[pmlr,twocolumn]{jmlr} 

\usepackage{booktabs}
\usepackage[load-configurations=version-1]{siunitx} 

\theorembodyfont{\upshape}
\theoremheaderfont{\scshape}
\theorempostheader{:}
\theoremsep{\newline}

\jmlrvolume{ML4H Extended Abstract Arxiv Index}
\jmlryear{2020}
\jmlrsubmitted{2020}
\jmlrpublished{}
\jmlrworkshop{Machine Learning for Health (ML4H) 2020}

\title[Towards Trainable Saliency Maps]{Towards Trainable Saliency Maps in Medical Imaging}

\author{%
\begin{center}
\Name{Mehak Aggarwal$^1$}, \Name{Nishanth Arun$^1$}, \Name{Sharut Gupta$^1$}, \Name{Ashwin Vaswani$^1$}, \Name{Bryan Chen$^1$}, \Name{Matthew Li$^1$}, \Name{Ken Chang$^{1,2}$}, \Name{Jay Patel$^{1,2}$}, \Name{Katherine Hoebel$^1$}, \Name{Mishka Gidwani$^1$}, \Name{Jayashree Kalpathy-Cramer$^1$}, \Name{Praveer Singh$^{1,2}$}
\\
\addr $^1$ Athinoula A. Martinos Center for Biomedical Imaging, Boston, MA, USA
 \\ $^2$ Massachusetts Institute of Technology, Cambridge, MA, USA
\end{center}
}

\begin{document}

\maketitle

\begin{abstract}
 While success of Deep Learning (DL) in automated diagnosis can be transformative to the medicinal practice especially for people with little or no access to doctors, its widespread acceptability is severely limited by inherent black-box decision making and unsafe failure modes. While saliency methods attempt to tackle this problem in non-medical contexts, their apriori explanations do not transfer well to medical usecases. With this study we validate a model design element agnostic to both architecture complexity and model task, and show how introducing this element gives an inherently self-explanatory model. We compare our results with state of the art non-trainable saliency maps on RSNA Pneumonia Dataset and demonstrate a much higher localization efficacy using our adopted technique. We also compare, with a fully supervised baseline and provide a reasonable alternative to it's high data labelling overhead. We further investigate the validity of our claims through qualitative evaluation from an expert reader.
\end{abstract}

\begin{table*}[htbp]
\floatconts
  {tab:operatornames}%
  {\caption{Classification AUC scores (top), Pointwise Detection scores (middle) and Localization Utility AUPRC scores with Retinanet baseline (bottom)}}%
  {\begin{tabular}{lllllllll}
  \toprule
  \bfseries Models & \bfseries R-18  & \bfseries R-34  & \bfseries R-50  & \bfseries R-101  & \bfseries R-152  & \bfseries D-121 & \bfseries D-169  & \bfseries V-19\\
  \midrule
  Baseline & 98.39 & 98.17 & 98.32 & 98.45 & 97.14 & 96.65 & 97.39 & 98.72\\
  Modified & 98.77 & \textbf{98.94} & \textbf{98.92} & \textbf{98.80} & \textbf{98.97} & \textbf{97.90} & \textbf{98.10} & \textbf{98.90}\\
  Retinanet & \textbf{98.81} & 98.16 & 97.08 & 97.88 & 98.25 & - & - & -\\
  
  \midrule
  Modified & 32.74 & 24.35 & 31.45 & 24.17 & 43.84 & 30.40 & 30.25 & 33.08\\
  Retinanet & 38.75 &  35.14 & 32.40 & 37.01 & 36.16 & - & - & -\\
  
  \midrule
  GRAD & 38.26 & 31.37 & 31.36 & 37.71 & 33.21 & 27.88 & 31.92 & 25.09\\
  SG & 26.05 & 29.21 & 30.41 & 36.27 & 31.55 & 29.76 & 30.34 & 18.55\\
  IG & 41.19 & 34.09 & 29.46 & 38.73 & 34.16 & 28.47 & 30.54 & 22.58\\
  SIG & 31.07 & 34.98 & 30.10 & 39.11 & 31.79 & 29.76 & 30.41 & 18.60\\
  GCAM & 32.83 & 45.91 & 48.66 & 45.71 & 24.67 & 29.57 & 37.27 & 24.69\\
  GBP & 38.26 & 31.37 & 31.36 & 37.71 & 33.21 & 27.88 & 31.92 & 28.27\\
  GGCAM & 44.76 & 39.55 & 39.49 & 45.22 & 32.30 & 34.66 & 37.05 & 32.80\\
  Modified & \textbf{48.04} & \textbf{46.18} & \textbf{48.16} & \textbf{48.43} & \textbf{50.27} & \textbf{42.87} & \textbf{45.56} & \textbf{54.84}\\
  
  \midrule
  
  Retinanet & 57.55 & 54.99 & 52.91 & 54.56 & 56.13 & - & - & -\\

  \bottomrule
  \end{tabular}}
\label{tab:tab1}
\end{table*}

\section{Introduction}
\label{sec:intro}
Image localization is a crucial supplement to image classification especially in clinical contexts as it allows 1) comprehension of model decision 2) assessment of failure modes and, 3) identification of regions of interest (ROIs) that can guide subsequent imaging and clinical management. Approaches offering fine localization rely heavily on detailed and time consuming annotations either as bounding boxes or pixel level segmentations from clinical experts. To limit the need for large-scale annotations, methods employing Semi Supervised Learning (SSL)\citep{chen2019multi} and Weakly Supervised Learning (WSL)\citep{lian2019end, wang2019weakly} have recently gained much attention from the medical imaging community. While the SSL framework augments a small pool of labeled samples with a large pool of unlabeled samples, WSL requires training with noisy or uncertain labels. Additionally, post-hoc saliency methods have long been used in a classification setup to explain classifier outputs \citep{mahapatra2016retinal, pasa2019efficient,bush2016lung} and as medical abnormality localizers\citep{amit2017hybrid, deepak2013visual}. However, since they are not trained end-to-end with the network, they offer limited efficacy in medical imaging applications \citep{arun2020assessing}. As more models are developed for routine usage by clinicians, interpretability and localizability of findings remain pivotal to building trust and acceptability in medical community.

\paragraph{Contribution}
In this study, we adapt a previously developed WSL technique to deliver loose bounding-box localization, utilizing only image-level classification labels for pneumonia in chest radiographs. More specifically, our method automatically outputs the ROI responsible for making pneumonia prediction in the given classification model without 1) need for retraining 2) loss of classification performance 3) ground truth annotations. We further compare our method with non-trainable saliency based methods using a quantitative and qualitative set of evaluation metrics for localization success and demonstrate a failure mode assessment of the model by a radiology resident.

\begin{figure*}[htbp]
\floatconts
  {fig:maps}
  {\caption{Respective saliency methods visualized with inverted heatmaps highlighting ROIs}}
  {\includegraphics[width=1\linewidth]{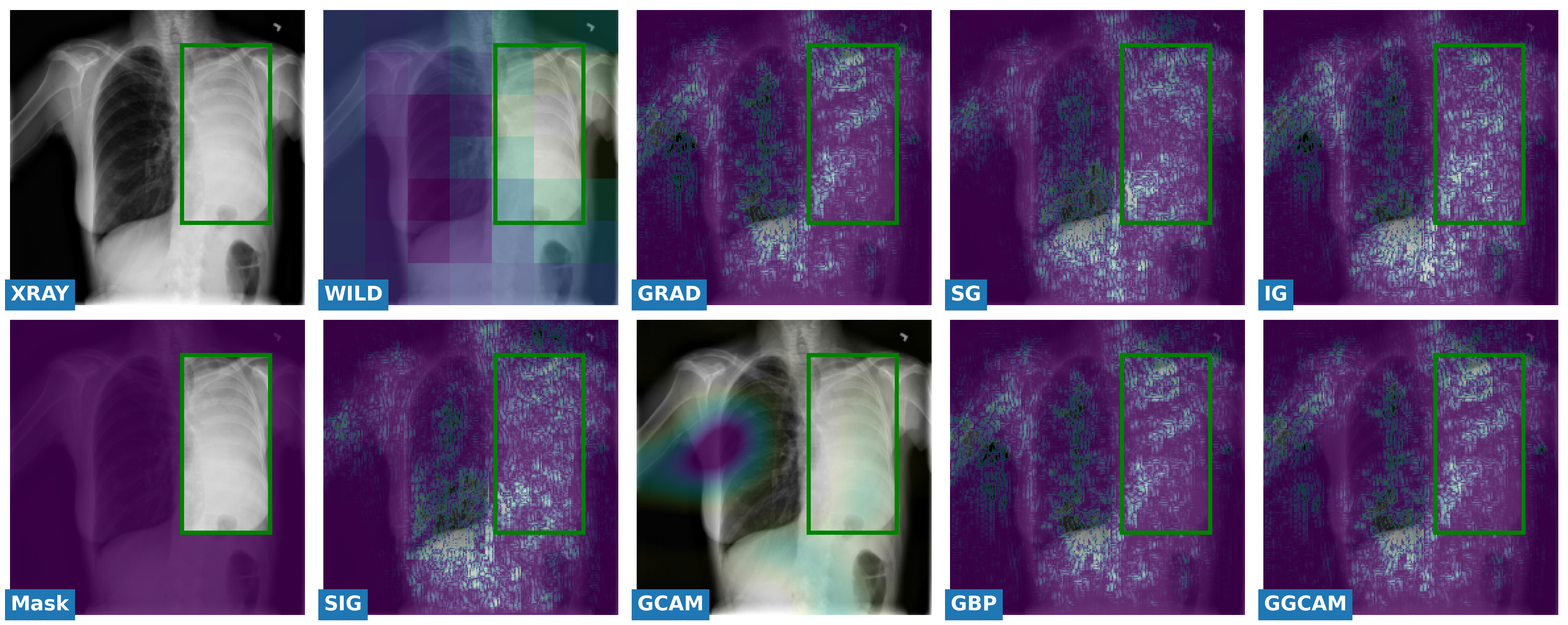}}
\end{figure*}

\section{Setup}
\label{sec:setup}


\paragraph{Dataset}
The dataset used for this analysis comprises of chest radiographs from the second stage of RSNA Pneumonia Detection Challenge \citep{RSNA} and contains $26,684$ scans, each labelled with one of 'Lung opacity', 'No Lung Opacity/Not Normal', and 'Normal'. For the purpose of this study, we excluded all 'No Lung Opacity/Not Normal' images, so as to restrict lung opacity as a pneumonia indicator. The dataset was then split in a 80:10:10 ratio with nearly 40\% positive pneumonia cases, all having bounding box annotations.

\paragraph{Architecture}
The most commonly used convolutional neural network (CNN) architectures for classification, such as Densenet \citep{iandola2014densenet}, Resnet \citep{resnet} and VGG \citep{vgg}, are primarily composed of two segments 1) Feature Network 2) Classifier Network. Of these, the former preserves spatially relevant context information, an element that is critical to the success of the WILDCAT model design \citep{durand2017wildcat} which we subsequently use. The latter, both here and in general, is often a fully connected network that outputs the prediction probability score.

In this technique, the given classifier network is replaced with two spatially aware pooling layers. The first layer splits the feature network output into Maps (\emph{m}) * Classes (\emph{c}) number of channels, which are then pooled to yield an activation map for each class, thus, forming the first part of the two valued output. These classwise maps are then fed into a spatial pooling layer which takes weighted (1:$\alpha$) average of the \emph{k} highest and lowest activation pixel values to give the final prediction score.


\paragraph{Experiments}
To validate generalizabilty of above technique, we present results over 8 different architectures and parameter counts - Resnet-18, 34, 50, 101, 152, Densenet-121, 169 and VGG-19. For comparison we also present results over 7 commonly used saliency method techniques - Gradient Explanation (GRAD), Smoothgrad (SG), Integrated Gradients (IG), Smoothgrad on Integrated Gradients (SIG), GradCAM (GCAM), Guided-backprop (GBP), and Guided GradCAM (GGCAM) along with a state of the art object detection framework, Retinanet, variants of which occupy top three positions on the original challenge leaderboard \citep{RSNA}. Each model presented has undergone multiple runs with different seed values, early stopping, minimum data pre-processing and an extensive hyperparameter search (Appendix \ref{apd:second}). 

\section{Results}
\label{sec:results}

\paragraph{Classification}


As shown in Table \ref{tab:tab1} (top block), modified architectures always outperform baseline for each of the model types considered, thus, verifying that we not only maintain classification performances but are better by 0.5-1\%. We also find the best results are consistently obtained at low \emph{k} values, implying the model focuses only on specific areas (ROIs) in the modified architecture to make a successful prediction.

\paragraph{Pointwise Detection}


For detection evaluation by our models, we follow the pointwise detection metric adopted by the Wildcat \citep{durand2017wildcat} paper. This metric captures the quality of detection without penalizing the lack of an appropriately sized bounding box. Each classwise map (\textit{s*s}) is extrapolated to match the size of the original input image (\textit{S*S}) resulting in fixed size bounding box (\textit{d*d}, where \textit{d = S/s}). Table \ref{tab:tab1} (middle block) reports mean Average Precision scores, with modified ResNet-152 performing better by a 5\% margin than the best performing Retinanet network, our fully supervised detection baseline.

\paragraph{Localization Utility}


To improve interpretabilty and reliability of classification tasks, it is imperative to provide clinicians with meaningful visualizations of model rationale behind a prediction. We thus aim to provide high quality visualizations with classwise maps. The extrapolated map can be viewed in juxtaposition with the original radiograph (first row, second column in \ref{fig:maps}) to highlight abnormal regions which are suspected pneumonia ROIs. To justify their efficacy, we compare classwise map, saliency methods' map and retinanet box output map with ground truth annotation mask, and calculate a simple similarity score (Appendix \ref{apd:second}) between former and latter to provide results. Classwise maps are clearly superior to the saliency maps. The best performing saliency method, GGCAM, is still worse off by 10\% on average than modified architecture and the best performing model type, VGG19, has comparable performance to average performance of fully supervised baseline.

\paragraph{Reader Study}


\begin{figure}[htbp]
\floatconts
  {fig:reader}
  {\caption{FN (Left) and FP (Right)}}
  {\includegraphics[width=0.7\linewidth]{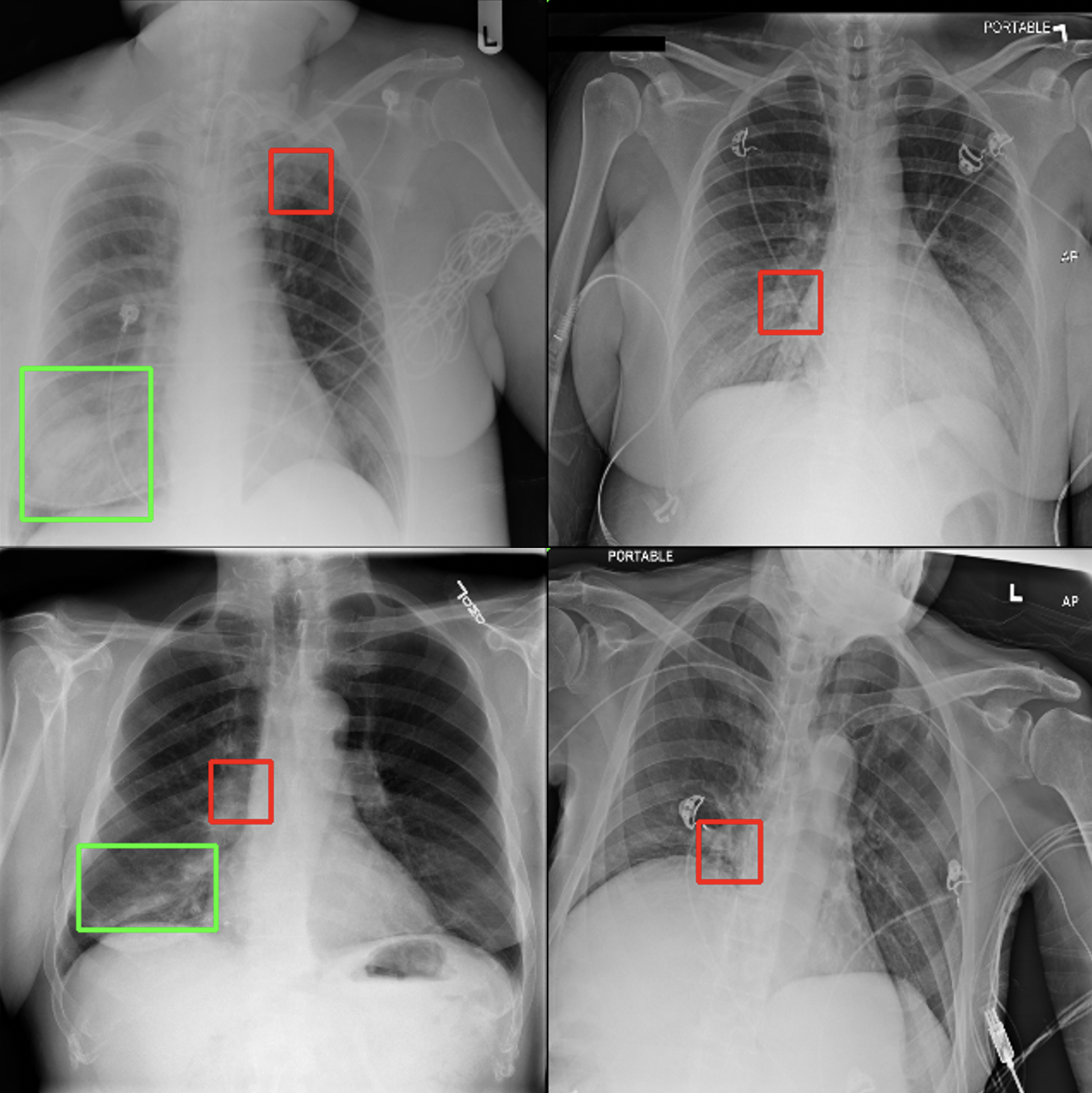}}
\end{figure}

A senior diagnostic radiology resident inspected model failure cases. The cases were divided into classification false positives (FP) and true positives but with no predicted box overlap with ground truth bounding box (FN). As seen in Fig \ref{fig:reader}, for FP cases, most showed localization to either the hilum of lung or a clavicle projecting over lung apex. Both sites represent sites of increased density that the model mistook for pneumonia. For FN cases, we similarly find that the predicted bounding box tends to localize to lung hilum (and to lesser extent the clavicle projecting over lung apex), instead of actual lung pathology of interest. Overall model tends to incorrectly identify normal opacities of the hilum and clavicle as pneumonia. These are challenging features for the model to learn, as many diverse pathologies occur at the hilum and lung apices but even here, clinical failure can be avoided by quick confirmation with the predicted box.

\section{Conclusion}
\label{sec:conclusion}

In this study, we validate a WSL pipeline for interpretation and abnormality localization for pneumonia classification models using only image-level labels. Our approach is trained end-to-end with classification labels, improving base performance and outperforming saliency map methods which are commonly used in medical literature for model interpretation and localization. Our results have implications for the clinical deployment of deep learning models, increasing their utility and explainability for clinicians. This may facilitate their trust in the models and expedite integration into the clinical workflow.

\bibliography{jmlr-sample}

\appendix

\section{Training Details}\label{apd:first}

For both the baseline classification model and the modified WILDCAT model, a batch size of 64 images, binary cross entropy loss function, Adam optimizer and early stopping with patience of 10 epochs on loss change was used across models. Best baseline model across mutiple learning rates and best WILDCAT model across mutiple learning rates,  \emph{k}, $\alpha$ and \emph{m} parameters was used to present results. Both localization utility and pointwise detection results are presented over best performing classification models, mimicking real world deployment where lack of ground truth annotations make optimizing over localization performance intractable. Overall, multiple runs with different seeds, amounting to over 1200+ trained models were analysed to effectively validate statistical reliability of each presented result.

\section{Metric Explanations}\label{apd:second}

\paragraph{Pointwise Detection} The metric identifies a successful localization as one where the center point of the predicted bounding box lies inside the ground truth annotation box. Similar to the standard mean Average Precision (mAP) calculation for object detection, the True positives defined here, should satisfy both high confidence score and the localization criteria given above. False positives and False negatives for the same can be defined congruently.

\paragraph{Localization Utility} The metric estimates the area under the precision recall curve (AUPRC) by treating, for single image, the binary array obtained from annotation mask as the ground truth labels and, the continuous, normalized array obtained from saliency maps as the predicted scores. From here, the calculated AUPRC score for each image is averaged to give the final localization utility result. Thus, the metric clearly captures extent of similarity between given annotation mask and the saliency map. In the figure {fig:auprc} given below, we present a sample saliency map and it's corresponding precision recall curve. Further, for Retinanet detection baseline, the map is obtained by allocating each pixel the value of the highest scoring box that covers it.

\begin{figure}[htbp]
\floatconts
  {fig:auprc}
  {\caption{Saliency map and its PR Curve}}
  {\includegraphics[width=\linewidth]{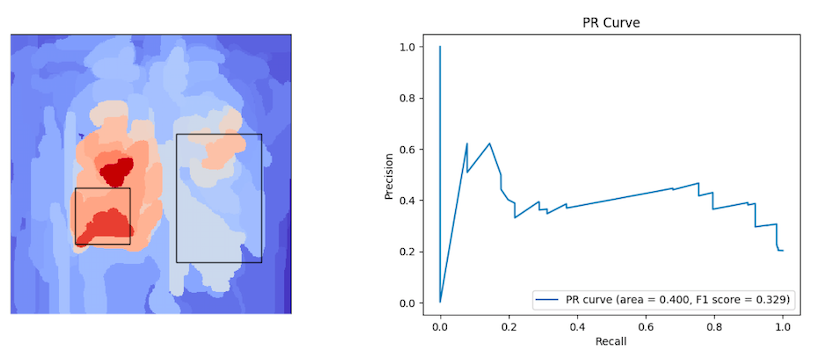}}
\end{figure}

\end{document}